\documentclass{amia}
\usepackage{amsmath}
\usepackage{amssymb}
\usepackage{booktabs}
\usepackage{xcolor}
\usepackage{multirow}
\usepackage[normalem]{ulem}
\usepackage{wrapfig}
\newcommand{\JakirTODO}[1]{\textcolor{red}{#1}}

\usepackage{url}
\usepackage{comment}
\excludecomment{Comment}

\begin{document}

\title{Knowledge Graph–Augmented Ambient AI for Clinical Note Generation}

\author{
Jakir Hossain, PhD$^1$,
Yi-Fei Zhao, BS$^{1,2}$,
Hongjian Wang, MS$^1$,
Minmei Shih, PhD, OTR/L$^3$,
Katie Leigh Mullen, OTR/L$^3$,
Ahmad P. Tafti, PhD$^4$,
Leming Zhou, PhD$^4$,
Manoj Purohit, MS$^5$,
William Hogan, MD$^5$,
Jay Zeng, MS$^1$,
Elizabeth Skidmore, PhD, OTR/L$^3$,
Yanshan Wang, PhD$^{1,6}$
}

\institutes{
    $^1$ Department of Biomedical Informatics, University of Pittsburgh, Pittsburgh, PA \\
    $^2$ Stern School of Business, New York University, New York, NY \\
    $^3$ Department of Occupational Therapy, University of Pittsburgh, Pittsburgh, PA \\
    $^4$ Department of Health Information Management, University of Pittsburgh, Pittsburgh, PA \\
    $^5$ Medical College of Wisconsin, Milwaukee, WI \\
    $^6$ Clinical and Translational Sciences Institute, University of Pittsburgh, Pittsburgh, PA
}

\maketitle

\section*{Abstract}
\textit{Ambient AI is increasingly adopted in healthcare to automatically generate clinical notes from patient–clinician conversations, with the potential to substantially reduce clinician documentation burden. However, generated notes may omit clinically relevant information discussed during the encounter, creating information gaps that can affect downstream care. Knowledge graphs (KGs) constructed from encounter transcripts can provide a structured representation of what was discussed and enable systematic identification of missing information from generated notes that are critical for patient care. In this study, we introduce Coverage-Directed Revision (CDR), a model-agnostic framework that constructs a KG from the encounter transcript, identifies medical concepts absent from an initially generated note, and directs large language models (LLMs) to restore the missing information without modifying the underlying note-generation system. We evaluate CDR on two datasets: 1) Pitt-Bench, a local dataset comprising rehabilitation sessions, and 2) ACI-Bench, a public dataset for benchmarking clinical note generation. We tested four underlying LLMs widely used in ambient AI systems. The results show that CDR consistently improves content recall across all evaluated conditions. Our study provides a practical approach for improving the completeness of ambient AI-generated clinical documentation.
}

\section*{Introduction}

Clinical documentation consumes a substantial portion of clinicians' time, motivating the growing adoption of ambient AI scribes that automatically generate clinical notes from recorded patient--clinician encounters\cite{lukac2025rct}. These notes are commonly structured in the SOAP format, which organizes information into what the patient reports (Subjective), what the clinician observes or measures (Objective), the clinician's interpretation of the patient's condition (Assessment), and subsequent actions or treatment decisions (Plan). Generating accurate and complete SOAP notes from clinical conversations, however, is challenging. Patient--clinician conversations are inherently unstructured, clinically relevant information may be distributed throughout the encounter, and clinician-authored reference notes are often concise and expressed in dense clinical shorthand. Although reasoning-enabled large language models (LLMs) have demonstrated strong performance on medical reasoning benchmarks\cite{bedi2025medhelm,wang2025medical}, these gains do not necessarily translate to clinical note generation. Recent evidence suggests that explicitly enabling reasoning may even degrade the quality of generated SOAP notes\cite{sourceaware2026}.

Evaluations of ambient AI systems have largely focused on whether generated notes introduce unsupported or fabricated content. However, clinically relevant information discussed during the encounter may be absent from the generated note\cite{asgari2025creola}. Such omissions can be difficult to detect during routine review, and information that never enters the clinical record becomes unavailable to downstream clinicians and other users of the chart\cite{automationbias2025}. Consequently, omission is increasingly recognized as a distinct failure mode, with dedicated metrics developed to quantify it\cite{schumacher2025medomit} and approaches proposed to mitigate it through retraining of the AI model\cite{claimrewards2025}. Retraining, however, may not be feasible for proprietary or already deployed ambient AI systems. We therefore focus on \textit{recall}: the proportion of clinically relevant encounter content recovered in the generated note. Accurately measuring recall presents an additional challenge. Conventional evaluations typically treat the clinician-written note as a complete reference, even though clinical notes are inherently selective summaries of the encounter. As a result, transcript-supported information that the clinician chose not to document may be incorrectly penalized as erroneous, while information documented by the clinician but never stated during the recorded encounter may be counted as a model omission. These mismatches confound failures of the note-generation system with differences between the conversation and the clinician-written reference note.

Knowledge graphs (KGs) are well suited to this problem because a graph constructed from the encounter transcript provides a structured representation of what was actually discussed. Existing approaches have primarily incorporated such graphs into the generation prompt to guide note generation\cite{kosmos2026}. An alternative is to use the graph post hoc where the KG can serve as a coverage representation against which an already generated draft is checked. This use of KGs has been explored in general summarization\cite{zhu2021fasum}, but remains underexplored in clinical note generation, where post-hoc revision methods typically operate without an explicit graph-based representation of source content\cite{gao2023rarr,madaan2023selfrefine}. The closest related approach iteratively re-summarizes a partial clinical note across multiple passes\cite{cadence2023}, but does not explicitly identify which source-supported content is missing from the draft.

We pursue this post-hoc approach and focus on enhancing the recall for the ambient AI systems. We introduce \textit{Coverage-Directed Revision (CDR)}, which constructs a KG from the encounter transcript, compares the graph against an initially generated note to identify concepts absent from the draft, and directs the  ambient AI model to incorporate the missing content. Crucially, the KG functions as a \textit{coverage check} rather than as input to the initial generation process. It therefore does not alter the base model or its prompt: the initial draft remains exactly what the underlying ambient AI model would have produced independently. This separation makes CDR model-agnostic and enables it to be applied as a post-hoc revision layer to existing, already deployed ambient AI systems.


This paper has three contributions. First, we introduce \textit{CDR}, a model-agnostic, post-hoc revision framework that can be applied to any base ambient AI model without modifying its generation process. CDR improves recall across every condition evaluated, with gains of up to 5.59 points under concept-overlap scoring and 7.08 points under atomic-claim scoring. 
Second, we introduce a transcript-grounded evaluation that stratifies recall according to whether each clinician-documented claim is entailed by the encounter transcript. Using majority labels from five independent judges, this evaluation distinguishes content omitted by the models from content that was never available to an audio-based system. Last, we evaluate CDR on Pitt-Bench, a real-world, multidisciplinary rehabilitation corpus, and on the publicly available ACI-Bench benchmark, demonstrating its effectiveness across datasets and clinical settings.

\section*{Background and Related Work}


\noindent\textbf{Ambient note generation and evaluation.} ACI-Bench was
introduced as a benchmark for generating clinical notes automatically from doctor--patient
conversation\cite{yim2023acibench}. Its baseline fine-tunes a seq2seq model and writes the note
one section at a time, which beats writing it whole. Later work prompts an LLM with retrieved
encounters as in-context examples\cite{summqa2023}, fine-tunes smaller models, or splits the
work across agents\cite{agenticsoap2026}. Benchmarks rate the results
well\cite{bedi2025medhelm}, though switching on reasoning can make SOAP notes
worse\cite{sourceaware2026}. Scoring has focused on invention\cite{asgari2025creola}, and
ROUGE\cite{lin2004rouge} and BERTScore\cite{zhang2020bertscore} measure how similar a generated
note is to the clinician note without saying which part of the reference is missing. Omission
is now a failure mode in its own right, with metrics built for it\cite{schumacher2025medomit},
and recent work splits into two families. Concept-based metrics compare the clinical concepts
in the two notes. MEDCON, introduced with ACI-Bench\cite{yim2023acibench}, extracts UMLS
concepts with QuickUMLS\cite{soldaini2016quickumls} and reports F1 over the two concept sets.
It keeps semantic groups like Anatomy, Device and Disorders, but excludes Procedures and
Activities, which are most of a therapy note, so MEDCON reads low on rehabilitation by
construction. Claim-based metrics work finer, splitting each note into atomic statements.
DocLens\cite{xie2024doclens} extracts claims from both texts and matches them, so an unmatched
claim is a hallucination or an omission\cite{maynez2020faithfulness}.
FactEHR\cite{munnangi2024factehr} adds a step DocLens lacks: it drops any claim its own note
fails to entail before comparison, so decomposition errors never reach the score. Recall is
then the share of clinician claims the generated note entails, precision the reverse. Both
families treat the clinician note as truth, which confuses two errors: content the model
invented, and content that is true of the visit but unwritten. Grounding claims and concepts
against the transcript separates them. We use both metric families, and analyse the
FactEHR claims against the transcript as well.

\noindent\textbf{Knowledge graphs and post-hoc revision.} A graph built from the transcript
records what was said, which makes it a natural handle. Existing systems use it during
generation. KOSMOS\cite{kosmos2026} is closest: it extracts typed entities, attributes and
relations, grounds them to UMLS under an ontology schema, keeps links to the supporting turns,
and gives the graph to the model, reporting the highest raw scores on ACI-Bench among its
conditions. DR.KNOWS instead retrieves UMLS paths for diagnosis summarisation from SOAP
notes\cite{gao2025drknows}, and graph retrieval augments generation over document
collections\cite{edge2024graphrag,hossain2026core}. In every case the graph enters the prompt.
Editing after generation is also established. Zhu et al.\ build a graph from the source article
and use it both to condition generation and to post-edit other systems'
summaries\cite{zhu2021fasum}. Without a graph, Self-Refine has a model critique and rewrite its
own output\cite{madaan2023selfrefine}, and RARR revises text so its statements become
attributable to retrieved evidence\cite{gao2023rarr}. All act on unsupported content: they
remove or correct what the evidence does not support. In clinical notes the nearest work
re-summarises a partial note over several passes\cite{cadence2023}, with no signal for what is
missing, while systems that do target omission retrain the
model\cite{claimrewards2025}. That leaves one gap: no prior work uses a transcript-derived
graph to direct the revision of a finished clinical note. The proposed CDR fills the gap, applying the graph
after generation as an audit of what the draft omitted and leaving the base model
untouched.



\section*{Methods}

An ambient AI system first generates a draft clinical note from the encounter transcript, but the resulting draft may omit clinically relevant content present in the source conversation. \textit{Coverage-Directed Revision (CDR)} operates as a post-hoc revision step to recover such omissions. CDR constructs a KG from the same transcript, compares the graph against the generated draft to identify concepts that are not represented, and directs the model to incorporate the missing content into the note. Importantly, CDR does not modify the underlying  ambient AI model or its generation process. It can therefore be applied as a modular revision layer to existing ambient AI systems, including those that are already deployed.

\subsection*{Overview}

\begin{wrapfigure}{r}{0.52\textwidth}
\vspace{-\intextsep} 
\centering
\includegraphics[width=0.5\textwidth]{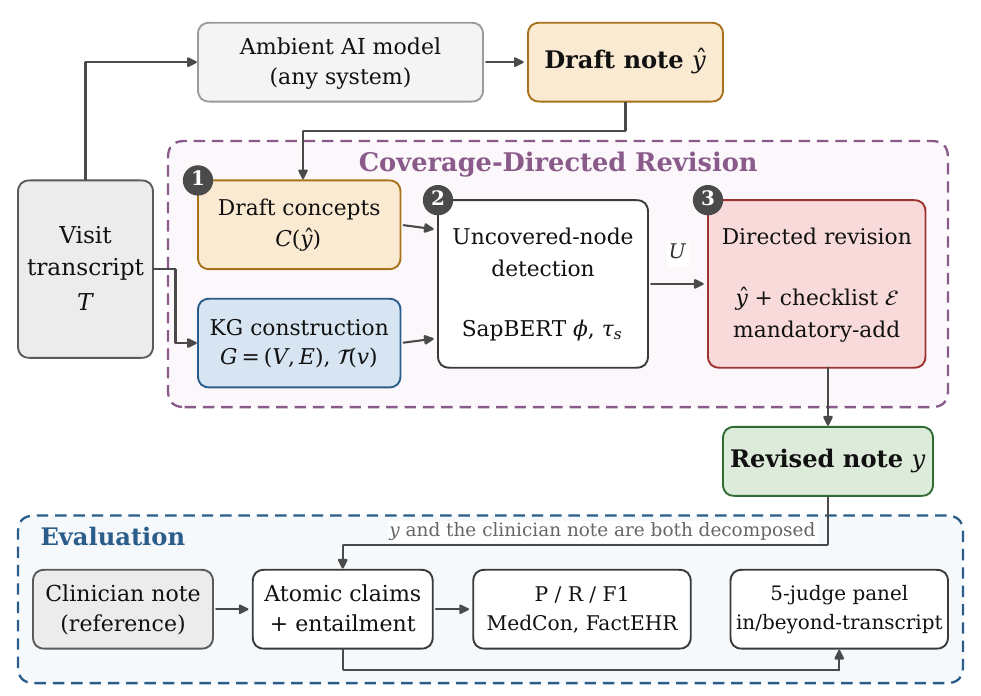}
\caption{An overview of the proposed knowledge graph-augmented ambient AI for clinical note generation. 
}
\label{fig:system}
\end{wrapfigure}

Figure~\ref{fig:system} provides an overview of the CDR framework. The revision is explicitly \emph{post-hoc}: the KG is never incorporated into the initial generation prompt, ensuring that the draft $\hat{y}$ is exactly what the underlying  ambient AI model would have produced independently. Consequently, any difference between the initial draft $\hat{y}$ and the revised note $y$ can be attributed to the CDR revision step rather than to changes in the base generation process. This separation also makes CDR portable across note-generation systems. The implementation requires no access to the internal architecture, parameters, or generation procedure of the base model, which is therefore represented as a black box in Figure~\ref{fig:system}.
The lower panel previews the evaluation, described in Evaluation Metrics later.


\subsection*{Setup and Notation}

Let $\hat{y}$ denote a draft SOAP note produced by a base model from a visit transcript
$T=\{t_1,\dots,t_M\}$, and let $G=(V,E)$ be the KOSMOS KG extracted from $T$. The
graph is built by prompting an LLM over the numbered turns of $T$ with the KOSMOS extraction
prompt, which returns typed entities, the relations between them, and the turns each was drawn
from. Its nodes $v\in V$ are coreference-grouped clinical concepts, each carrying a canonical
name $\mathrm{name}(v)$, an entity type $\tau(v)$, and the set of transcript turns
$\mathcal{T}(v)\subseteq T$ in which it is mentioned. CDR uses the nodes and discards the relations for simplicity. We denote the revised note by $y$ and refer to the model responsible for producing $y$ as the \emph{reviser}. Importantly, the \emph{reviser} is not required to be the model that produced the draft. We evaluate both same-model and cross-model revision settings to assess whether CDR's effectiveness depends on the choice of \emph{reviser}.

\subsection*{Coverage-Directed Revision}
\label{sec:cdr}

  CDR treats $G$ as an \emph{omission signal}: rather than regenerating $\hat{y}$, it
  identifies concepts that are supported by the transcript and present in $G$ but absent from
  the draft, and revises $\hat{y}$ to insert them. It consists of three stages.

  \paragraph{(1) Draft concept extraction.}
  First, we prompt the \emph{reviser} to enumerate the clinical concepts already
  expressed in the draft, yielding a set $C(\hat{y})=\{c_1,\dots,c_n\}$. This makes coverage
  assessment operate at the concept level rather than on surface strings, so that paraphrases in
  $\hat{y}$ are credited as covered.
We use the \emph{reviser} for this step, rather
than a separate extractor, so that the concepts in $C(\hat{y})$ are named in the same words the \emph{reviser} will use when it rewrites the note. All calls use greedy decoding. 
If extraction returns nothing, $C(\hat{y})=\varnothing$, coverage then depends on exact string matching alone, which marks more nodes as missing than it should.

  \paragraph{(2) Uncovered-node detection.}
  A KG node is deemed \emph{uncovered} if it is neither lexically nor semantically present
  in the draft. Nodes are first deduplicated by canonical name, preserving graph order. Using a
  frozen SapBERT encoder $\phi(\cdot)$\footnote{\url{https://huggingface.co/cambridgeltl/SapBERT-from-PubMedBERT-fulltext}}
  (run on CPU) and a similarity threshold $\tau_s=0.80$, the uncovered set is
  \begin{equation}
  U \;=\; \Big\{\, v \in V \;:\; \mathrm{name}(v)\notin \hat{y}
  \;\wedge\; \max_{c\in C(\hat{y})} \cos\!\big(\phi(\mathrm{name}(v)),\,\phi(c)\big) < \tau_s \,\Big\}\,,
  \end{equation}
  i.e.\ nodes whose canonical name neither appears verbatim in the note nor matches any
  extracted draft concept above $\tau_s$. $U$ is used as extracted, with no further filtering,
  so the graph is judged as the pipeline produced it.

  \paragraph{(3) Directed, evidence-grounded revision.}
  For each node $v\in U$ we construct a \emph{directed checklist
  item} that (i) names the missing concept explicitly and (ii) grounds it in its own
  supporting turns:
  \begin{equation}
  \begin{aligned}[t]
   e(v)\;=\;\texttt{MISSING:}\;\text{``}\mathrm{name}(v)\text{''}\quad\texttt{SUPPORTED BY:}\;\mathcal{T}_2(v)
  \end{aligned}
  \end{equation}
  where the typewriter tokens denote fixed strings in the prompt and $\mathrm{name}(v)$, $\mathcal{T}_2(v)$ are
  substituted per node. $\mathcal{T}_2(v)\subseteq\mathcal{T}(v)$ is the first two mention turns
  in transcript order, quoted verbatim with their turn indices: two turns are enough to license
  an insertion, and capping keeps the evidence block short on dense graphs. A node contributing
  no resolvable turn is dropped from the block. The evidence block is their concatenation over the retained nodes,
  \begin{equation}
  \mathcal{E}(\hat{y},G)\;=\;\big(\,e(v)\,\big)_{v\in U}\,,
  \end{equation}
  which is passed with $\hat{y}$ to the \emph{reviser} under a \emph{mandatory-add} instruction: for every
  listed item whose quoted turn supports it, add a short factual sentence that explicitly
  names the concept, placed in the correct SOAP section, citing the supporting turns, and
  without altering unrelated content or returning the note unchanged. The \emph{reviser} emits the
  revised note $y$. Citations are requested so that each insertion stays anchored to a turn, but
  they are stripped from $y$ before scoring, so no metric rewards or penalises them. If
  $U$ is empty, the graph found nothing the draft omits and it is returned as it stands. One pass therefore costs one graph build plus two calls to
  the \emph{reviser}, and nothing is fine-tuned.
  Two features of this construction carry the method. Naming the concept, rather than leaving
  the \emph{reviser} to infer it from raw dialogue, both directs the insertion and encourages the
  lexicalization that concept-level scoring will match on. Requiring every supported item to be
  addressed prevents the near-identity revision a softer instruction invites, where the \emph{reviser}
  returns a note byte-similar to the draft.


\section*{Experimental Setup and Results}
 We first describe the dataset, prompts and metrics, then report results. Unless ACI-Bench is named, every number below is on Pitt-Bench.

\subsection*{Dataset}
We evaluate CDR on both private and public datasets. One is Pitt-Bench, a private
rehabilitation corpus of full-length therapy sessions with recordings, transcripts, and clinician-written gold standard clinical notes. The other is ACI-Bench, a public benchmark of mostly primary care outpatient transcripts and notes for validating ambient AI systems. 
Testing both shows whether the method holds across different datasets and clinical settings.

\begin{itemize}\itemsep0pt
  \item \textbf{Pitt-Bench.}  
  We evaluated on 45 adult inpatient rehabilitation
        therapy sessions from a single site, collected under IRB approval (STUDY25080111). The
        set covers three disciplines: 19 occupational therapy, 19 physical therapy, and 7
        speech-language pathology sessions. Each session was audio-recorded and transcribed with
        speaker diarization. Transcripts run 541 to 4{,}731 words (mean 2{,}189; median
        2{,}096). Each one is paired with the note the treating therapist wrote as part of
        routine documentation. Those notes are short: 67 to 398 words (mean 163; median 140). A
        note therefore keeps about one word for every thirteen spoken in the session. The data
        contain protected health information and cannot be shared.

  \item \textbf{ACI-Bench.} We used Test Set 1 of ACI-Bench ($n=40$), the MEDIQA-Chat 2023 Task B test split.
        It is the largest public dialogue-to-note benchmark: 207 simulated doctor--patient
        encounters (67 train, 20 validation, and three test sets of 40), each paired with an
        expert-written note in a fixed section structure. The encounters are general-medicine
        consultations, and they compress far less than ours --- about 1{,}230 tokens of dialogue
        to a 490-token note, or 2.5:1, against 13:1 in Pitt-Bench. Since ACI-Bench is also
        public, we compared our results with previous published work.
\end{itemize}


\subsection*{Large Language Models Tested as Ambient AI}

We evaluate four LLMs as ambient AI models: Qwen2.5-32B, Qwen2.5-72B, Qwen3-235B and GPT-5 Pro. Each produces a
draft note from the transcript, and CDR then revises that draft. Results are reported over two
draft sources. Our own SOAP prompt is the primary base note throughout. We add the KOSMOS
note-generation prompt as a second draft base, since KOSMOS is the state of the art on
ACI-Bench, to show that the gain is consistent on a draft we did not write. The two prompts
differ in more than wording, and Table~\ref{tab:prompts} sets out where.

\begin{table}[h]
\centering
\caption{The two draft prompts. Both ask for a SOAP note; they differ in context, section layout and citation requirements.}
\label{tab:prompts}
\footnotesize
\setlength{\tabcolsep}{4pt}
\renewcommand{\arraystretch}{1.15}
\begin{tabular}{@{}lp{0.43\textwidth}p{0.395\textwidth}@{}}
\toprule
 & \textbf{SOAP prompt} & \textbf{KOSMOS transcript-only condition prompt} \\
\midrule
Role
 & Ours, and the primary baseline. The corpus notes (PT, OT, SLP) are narrative SOAP records
   organised around function, not chief complaint and physical exam, and this is the prompt
   they were written under, so baselines compare against that reference rather than a
   re-specified target.
 & A second draft base, not our own: the KOSMOS note-generation prompt run on our transcripts with no graph supplied, so the gain is measured on a draft we did not write. KOSMOS adapts it from the DocLens SOAP
   prompt\cite{xie2024doclens}; we reuse it here for reproduction. \\
Context
 & Zero-shot: one system message followed by the transcript, with no exemplars. Being zero-shot it transfers across base models without in-context material that could leak into the output.
 & Two-shot: complete ACI-Bench primary-care encounters and notes as exemplars, about $20$k
   characters of in-context material carried over from the source template. \\
Sections
 & Narrative, paragraph-style SOAP, with each section enumerated: symptoms, concerns and
   presentation (Subjective); vital signs, assessment results and performed activity
   (Objective); interpretation of both (Assessment); goals, patient education and session
   planning (Plan).
 & Also asks for SOAP, so format is not the difference. But the layout the model \emph{sees}
   is ACI-Bench's: \textsc{History of Present Illness}, \textsc{Physical Exam},
   \textsc{Results}, \textsc{Assessment and Plan}, plus \textsc{Chief Complaint},
   \textsc{Review of Systems} and \textsc{Vitals Reviewed}. \\
Citations and graph
 & Neither is mentioned: no citation markers are required, and the prompt says nothing about a
   KG.
 & Both are: per-sentence citations to transcript turn indices, and a system message announcing
   a transcript-derived graph, which is empty in this condition. \\
\bottomrule
\end{tabular}
\end{table}

The KOSMOS prompt was written for primary-care encounters, and applying it unchanged to rehabilitation sessions made Qwen3-235B decline outright on 28 of 45, stating that the transcript was not a valid clinical encounter. 
We therefore append one paragraph naming the domain and forbidding refusal, leaving every other part of the prompt identical to the KOSMOS prompt.



\subsection*{Evaluation Metrics}


As described in Background, we use two metrics: MedCon measures overlap between the clinical
concepts in the two notes, and FactEHR measures entailment between their atomic claims. 
MedCon is scored at UMLS 2022 throughout, the release used when MedCon was introduced with
ACI-Bench\cite{yim2023acibench}. Concept counts shift with the release, so fixing one version
across every condition is what makes the numbers comparable between base models, corpora and
tables. MedCon is otherwise fast and deterministic.
Every condition we report (an ambient AI model
paired with a draft prompt on one corpus) is scored on both metrics, so no result rests on a
single metric. FactEHR uses a local Qwen2.5-72B judge, chosen from a five-model panel (Qwen2.5-32B, Qwen2.5-72B, Qwen3-235B, gpt-oss-20B, Llama-3.1-8B): it agrees closely with the larger Qwen3-235B at lower cost, and its self-verification keeps $99.8$–$99.9\%$ of claims.

Both metrics provide precision and recall in the same shape, but the two directions are not
symmetric. Recall is scored over the clinician note's content (under FactEHR, a mean of $19.1$
claims per note); precision is scored over the generated note's content (a mean of $36.9$).
We prioritize recall for two reasons. First, omissions of clinically relevant information from the encounter can have direct consequences for downstream patient care, care coordination, and reimbursement. Moreover, omissions appear to be more prevalent than hallucinations in ambient clinical documentation. In a recent evaluation of an ambient AI system, clinician review of nearly $13{,}000$ generated summary sentences identified an omission rate of $3.45\%$, compared with a hallucination rate of $1.47\%$\cite{asgari2025creola}. These findings suggest that incomplete documentation represents a substantial and potentially underappreciated source of error in ambient AI-generated notes. Second, omissions impose a distinct cognitive burden on clinician review. An incorrect or redundant statement that appears in a generated note can be directly identified and removed, whereas detecting an omission requires the clinician to recognize that information is absent and reconstruct the relevant details of the encounter from memory. This effectively reintroduces part of the documentation burden that ambient AI systems are intended to reduce. The problem may be further compounded by automation bias: clinicians reviewing AI-generated documentation may be less likely to detect information that is silently absent than information that is visibly incorrect\cite{automationbias2025}. Together, the downstream consequences of missing information and the difficulty of detecting omissions motivate our emphasis on recall as the primary measure of documentation coverage.

\noindent\textbf{Transcript grounding.} A clinician claim can be missing from the AI note
because the model dropped content that was in the recording, or because the content was never
in the recording at all. One recall score cannot tell these apart, and only the first is a model
failure. We therefore label every clinician claim by whether the transcript supports it, and
report recall separately for each group.
A claim the transcript entails is \emph{in-transcript}. A claim it does not entail is
\emph{beyond-transcript}: the therapist wrote it from watching and handling the patient, from
measurement, from the chart, or from clinical judgement. In rehabilitation this second group is
large. Gait quality, walking distance, assistance level and transfer performance are seen, not
said. CDR builds its checklist from the transcript alone, so it cannot reach that content.
Since a panel is more reliable than a single judge\cite{verga2024poll}, we label each claim with all five models above. Each reads the full transcript as
premise and one claim as hypothesis, answering yes or no; a claim is in-transcript when at least
three of the readable votes say yes (\emph{majority label}). We also consider the
\emph{unanimous subset}, where all five agree.

\subsection*{Results}


Table~\ref{tab:main_recall} reports recall for each base model before and after revision, under
both metrics: \emph{Draft} is the note the base model wrote, $+$CDR is that same note after CDR
has revised it, and $\Delta$ is the change between them. 
Each row's KG is built by its own base
model, except the two GPT-5 Pro rows, which reuse the Qwen3-235B KG due to the prohibitive cost of building one from full transcripts. Revision is self-revision
except in the indented row, where Qwen3-235B revises the GPT-5 Pro draft: the one cross-model
pair, included to test whether a local open-weight model can revise a frontier draft. Qwen2.5-32B is
omitted from the ACI-Bench SOAP block, where its drafts were degenerate, and GPT-5 Pro is
reported on Pitt-Bench only, on API-cost grounds.

\begin{table}[h]
\centering
\caption{Results of MedCon and FactEHR recall (\%) on Pitt-Bench ($n=45$) and ACI-Bench ($n=40$). 
}
\label{tab:main_recall}
\small
\setlength{\tabcolsep}{5pt}
\begin{tabular}{lcccccc}
\toprule
 & \multicolumn{3}{c}{\textbf{MedCon}} & \multicolumn{3}{c}{\textbf{FactEHR}} \\
\cmidrule(lr){2-4}\cmidrule(lr){5-7}
\textbf{Model} & Draft & $+$CDR & $\Delta$ & Draft & $+$CDR & $\Delta$ \\
\midrule
\multicolumn{7}{l}{\textit{\textbf{Pitt-Bench}, SOAP-prompt base}}\\
Qwen2.5-32B                      & 35.68 & 37.88 & $+2.20$ & 24.84 & 29.55 & $+4.71$ \\
Qwen2.5-72B                      & 37.57 & 42.95 & $+5.38$ & 27.13 & 34.21 & $\mathbf{+7.08}$ \\
Qwen3-235B                       & 45.67 & 50.22 & $+4.55$ & 38.93 & 41.56 & $+2.63$ \\
GPT-5 Pro                        & 48.59 & 51.26 & $+2.67$ & 43.37 & \textbf{46.61} & $+3.24$ \\
\quad revised by Qwen3-235B      & 48.59 & \textbf{52.94} & $+4.35$ & 43.37 & 45.09 & $+1.72$ \\
\midrule
\multicolumn{7}{l}{\textit{\textbf{Pitt-Bench}, KOSMOS transcript-only condition}}\\
Qwen2.5-32B                      & 19.48 & 23.11 & $+3.63$ & 16.67 & 19.20 & $+2.53$ \\
Qwen2.5-72B                      & 24.78 & 28.62 & $+3.84$ & 20.49 & 24.33 & $+3.84$ \\
Qwen3-235B                       & 34.88 & 40.47 & $+5.59$ & 26.80 & 32.56 & $+5.76$ \\
\midrule
\multicolumn{7}{l}{\textit{\textbf{ACI-Bench}, SOAP-prompt base}}\\
Qwen2.5-72B                      & 58.35 & 61.98 & $+3.63$ & 76.84 & 81.90 & $+5.06$ \\
Qwen3-235B                       & 60.42 & 65.45 & $+5.03$ & 82.89 & 84.91 & $+2.02$ \\
\midrule
\multicolumn{7}{l}{\textit{\textbf{ACI-Bench}, KOSMOS transcript-only condition}}\\
Qwen2.5-32B                      & 61.51 & 65.54 & $+4.03$ & 76.94 & 81.68 & $+4.74$ \\
Qwen2.5-72B                      & 66.23 & 68.42 & $+2.19$ & 78.35 & 84.18 & $+5.83$ \\
Qwen3-235B                       & 68.08 & 71.54 & $+3.46$ & 83.46 & 86.10 & $+2.64$ \\
\bottomrule
\end{tabular}
\end{table}

On Pitt-Bench, revision raises recall in all eight base $\times$ \emph{reviser}
combinations under both metrics: by $+2.20$ to $+5.59$ points on MedCon concept overlap and
$+1.72$ to $+7.08$ on FactEHR atomic claims. Concept counting and claim entailment measure
different units, so their agreement on direction in every cell is the substantive result; the
magnitudes are not expected to match.

The revision lifts every system without reordering them. Under both metrics the SOAP-prompt
base models finish in the order they started --- Qwen2.5-32B lowest, GPT-5 Pro highest --- so a
coverage pass complements a strong base model instead of substituting for one. The same holds
across bases: every KOSMOS transcript-only note starts far below its SOAP counterpart and stays
below it after revision.

A local \emph{reviser} is competitive with a frontier one. Handing the GPT-5 Pro draft to Qwen3-235B
gives the higher MedCon recall of the two ($52.94$ against $51.26$), while GPT-5 Pro
self-revision gives the higher claim recall ($46.61$ against $45.09$). Neither ordering is
consistent across metrics, so the defensible claim is the weaker one: the revision step does
not require a frontier API model and can run on local, PHI-safe infrastructure.

The trend persists on ACI-Bench. Recall rises in every condition under both metrics: $+2.19$ to
$+5.03$ points on MedCon and $+2.02$ to $+5.83$ on FactEHR. What differs is the starting point.
FactEHR draft recall on ACI-Bench runs
$76.8$--$83.5$ against $16.7$--$43.4$ on Pitt-Bench, so the ACI drafts already recover
most of what the reference note contains. That is what the corpus difference predicts: a
$2.5{:}1$ encounter leaves far less for the note to omit than a $13{:}1$ one, and an ACI-Bench
note is largely transcribed from speech, so little of it sits beyond the transcript. Less is
missing, so there is less for a coverage pass to find --- and the gains CDR does make are
therefore a larger share of the smaller gap, not a smaller effect.


\subsection*{KG-guided CDR Adds Grounded Information}
\begin{figure}[!tb]
\centering
\begin{minipage}{0.42\textwidth}
  \centering
  \includegraphics[width=\textwidth]{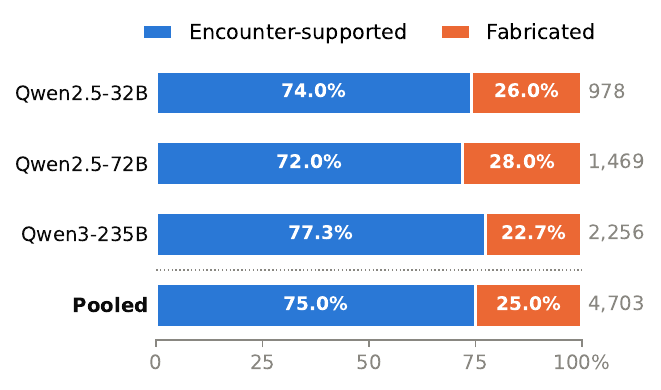}
\end{minipage}\hspace{0.04\textwidth}
\begin{minipage}{0.42\textwidth}
  \centering
  \includegraphics[width=\textwidth]{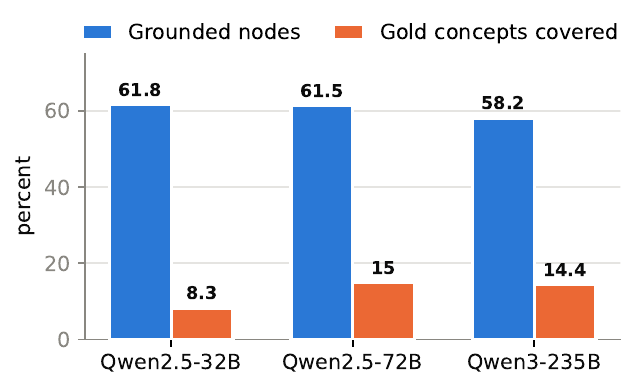}
\end{minipage}
\caption{Both panels are Pitt-Bench ($n=45$).  \textbf{Left:} the claims CDR
adds, grounded against the transcript by the Qwen2.5-72B judge; the count beside each bar is
the number of added claims.
\textbf{Right:}  KG quality. Grounded nodes is the share of graph nodes that map to a UMLS concept; Gold concepts covered is the share of the 648 gold-note concepts that appear in the graph.
}
\label{fig:kg_claims}
\end{figure}

CDR adds claims the clinician note does not contain, so precision scored against that note
falls, which may look like fabrication. 
But that note is
a selective summary, so a claim it omits may still be supported by the conversation. We
therefore ground each newly added claim against the transcript: one the transcript entails but
the clinician note omits is \emph{encounter-supported}; one that neither supports is
\emph{fabricated}.
We apply this grounding analysis to the content CDR introduces: the verified claims in the revised note, minus those already in the draft. Across the three base models, CDR adds 4703 claims, of which $75.0\%$ are entailed by the transcript (Figure~\ref{fig:kg_claims}, left). The revision is therefore mostly recovering content that occurred in the encounter but was left out of the written record, so the precision decline is largely a change in documentation scope rather than a rise in invention.
The rate holds across base models ($72.0$--$77.3\%$) despite very different volumes of added text ($21.7$ to $50.1$
claims per note), which suggests it is a property of the coverage mechanism rather than of any
one base model. The remaining quarter is either the honest cost of the method or a single judge
failing to support a claim against a long transcript, and we cannot separate the two here.

\noindent\textbf{The graph is a coverage signal, not a filter.} One potential strategy for recovering the precision lost during revision is to remove newly added claims that cannot be supported by the KG. However, Figure~\ref{fig:kg_claims} (right) illustrates why such filtering is not yet reliable. Only $58$--$62\%$ of graph nodes can be grounded to UMLS concepts, and each graph captures just $8$--$15\%$ of the gold-standard concepts present in its corresponding encounter. Thus, although the KG provides a useful signal for identifying content that may be missing from a draft, it does not constitute a sufficiently complete representation of the encounter to serve as a reliable filter for revised content. In particular, treating the absence of a claim from the graph as evidence that the claim is unsupported would risk removing valid, encounter-supported information. Improving KG extraction and strengthening the grounding between graph concepts and generated claims may ultimately enable such precision-oriented filtering; we leave this direction to future work.

\begin{Comment}

\begin{table}[h]
\centering
\caption{FactEHR claim recall split by whether the gold claim is entailed by the session
transcript. Pitt-Bench, $n=45$, Qwen2.5-72B model under self-revision; 863 gold claims
decomposed by Qwen2.5-72B. Each row uses a different source for the in-transcript label, so the
$n$ columns repartition the same claim set rather than describing different data; the consensus
row is the label used everywhere else in the paper. \emph{in-tx rate} is the share of claims
that label source counts as entailed, $\kappa$ is agreement with the Qwen2.5-72B labels, and
$p$-values are Wilcoxon signed-rank over encounters. The panel is entirely local, so no note or
transcript text leaves institutional infrastructure.
\JakirTODO{Equivalent tables for the Qwen2.5-32B and Qwen3-235B models have been computed and can
be added as a supplement; the in-transcript gain is significant under consensus for all three
($+7.0$, $+9.7$, $+4.7$).} }
\label{tab:stratified}
\scriptsize
\setlength{\tabcolsep}{3pt}
\begin{tabular}{lcccccccccc}
\toprule
 & & & \multicolumn{4}{c}{\textbf{in-transcript}} & \multicolumn{4}{c}{\textbf{beyond-transcript}} \\
\cmidrule(lr){4-7}\cmidrule(lr){8-11}
\textbf{Label source} & \textbf{in-tx rate} & \textbf{$\kappa$} & $n$ & Base R & Rev.\ R & $\Delta$ ($p$) & $n$ & Base R & Rev.\ R & $\Delta$ ($p$) \\
\midrule
\multicolumn{11}{l}{\emph{Generator: Qwen2.5-72B}}\\
Qwen2.5-72B  & 54.5\% & --    & 447 & 44.3 & 53.0 & $+8.7$ (0.000) & 372 & 8.1 & 9.7 & $+1.6$ (0.146) \\
Qwen3-235B   & 53.4\% & 0.685 & 437 & 45.3 & 54.7 & $+9.4$ (0.000) & 382 & 7.9 & 8.9 & $+1.0$ (0.480) \\
Llama-3.1-8B & 58.5\% & 0.419 & 482 & 38.4 & 46.3 & $+7.9$ (0.000) & 337 & 12.8 & 14.8 & $+2.1$ (0.017) \\
Qwen2.5-32B  & 35.7\% & 0.562 & 297 & 53.5 & 62.6 & $+9.1$ (0.000) & 522 & 13.2 & 16.7 & $+3.4$ (0.003) \\
gpt-oss-20B  & 23.5\% & 0.361 & 194 & 60.3 & 68.6 & $+8.2$ (0.010) & 623 & 17.7 & 22.3 & $+4.7$ (0.000) \\
\textbf{Majority label (3 of 5)} & \textbf{44.8\%} & -- & \textbf{373} & \textbf{50.4} &
\textbf{60.1} & $\mathbf{+9.7}$ \textbf{(0.000)} & \textbf{446} & \textbf{9.0} &
\textbf{11.0} & $+2.0$ (0.064) \\
\quad unanimous subset (5 of 5) & 40.9\% & -- & 149 & 66.4 & 75.2 & $+8.7$ (0.014) & 214 & 5.6 & 5.6 & $+0.0$ (1.000) \\
\midrule
\multicolumn{11}{l}{\emph{Generator: Qwen2.5-32B}}\\
Qwen2.5-72B  & 54.5\% & --    & 470 & 39.8 & 46.2 & $+6.4$ (0.000) & 392 & 6.9 & 8.7 & $+1.8$ (0.018) \\
Qwen3-235B   & 53.4\% & 0.685 & 461 & 40.6 & 47.3 & $+6.7$ (0.000) & 401 & 6.7 & 8.2 & $+1.5$ (0.058) \\
Llama-3.1-8B & 58.5\% & 0.419 & 505 & 35.2 & 40.2 & $+5.0$ (0.001) & 357 & 10.1 & 13.4 & $+3.4$ (0.003) \\
Qwen2.5-32B  & 35.7\% & 0.562 & 308 & 48.7 & 56.2 & $+7.5$ (0.001) & 554 & 11.6 & 14.1 & $+2.5$ (0.008) \\
gpt-oss-20B  & 23.5\% & 0.361 & 202 & 55.4 & 62.9 & $+7.4$ (0.005) & 658 & 15.3 & 18.7 & $+3.3$ (0.000) \\
\textbf{Majority label (3 of 5)} & \textbf{44.8\%} & -- & \textbf{387} & \textbf{45.0} &
\textbf{51.9} & $\mathbf{+7.0}$ \textbf{(0.000)} & \textbf{475} & \textbf{8.4} &
\textbf{10.5} & $+2.1$ (0.010) \\
\quad unanimous subset (5 of 5) & 40.9\% & -- & 156 & 61.5 & 67.9 & $+6.4$ (0.017) & 225 & 2.7 & 3.1 & $+0.4$ (0.317) \\   
\midrule
\multicolumn{11}{l}{\emph{Generator: Qwen3-235B}}\\
Qwen2.5-72B  & 54.5\% & --    & 470 & 57.9 & 61.5 & $+3.6$ (0.057) & 392 & 14.8 & 16.1 & $+1.3$ (0.214) \\
Qwen3-235B   & 53.4\% & 0.685 & 461 & 60.7 & 64.6 & $+3.9$ (0.048) & 401 & 12.5 & 13.5 & $+1.0$ (0.509) \\
Llama-3.1-8B & 58.5\% & 0.419 & 505 & 52.7 & 56.0 & $+3.4$ (0.071) & 357 & 17.9 & 19.3 & $+1.4$ (0.600) \\
Qwen2.5-32B  & 35.7\% & 0.562 & 308 & 66.9 & 71.1 & $+4.2$ (0.049) & 554 & 22.4 & 24.0 & $+1.6$ (0.289) \\
gpt-oss-20B  & 23.5\% & 0.361 & 202 & 72.8 & 77.7 & $+5.0$ (0.052) & 658 & 27.7 & 29.5 & $+1.8$ (0.162) \\
\textbf{Majority label (3 of 5)} & \textbf{44.8\%} & -- & \textbf{387} & \textbf{64.1} &
\textbf{68.7} & $\mathbf{+4.7}$ \textbf{(0.022)} & \textbf{475} & \textbf{17.3} &
\textbf{18.1} & $+0.8$ (0.417) \\
\quad unanimous subset (5 of 5) & 40.9\% & -- & 156 & 80.8 & 84.6 & $+3.8$ (0.113) & 225 & 7.6 & 8.9 & $+1.3$ (0.916) \\   
\bottomrule
\end{tabular}
\end{table}

\noindent\textbf{Finding.} The gain sits where it should. In-transcript claims improve by 4.5
points and the change is significant, while beyond-transcript claims do not improve
significantly. The KG is built only from the transcript, so it can only recover
things that were said; this is the pattern the method predicts.

The per-judge rows show the result does not hinge on one judge. All five give an in-transcript
gain between $+4.5$ and $+6.4$ points. The beyond-transcript gain is significant only under the
two smallest judges, which are also the two that agree least with the rest, and it disappears
when we keep only the claims all five judges agree on. What the judges do disagree about is
where the line sits: the share of claims counted as in-transcript ranges from 24\% to 57\%, so
we report a panel rather than a single number.

\end{Comment}

\subsection*{In-Transcript vs. Beyond-Transcript Recall}
If CDR works for the reason we claim, the gain should show up on in-transcript claims and not on
the rest.
Table~\ref{tab:stratified_consensus} splits recall by whether the transcript entails the claim, under both the majority label and the unanimous subset. The in-transcript gain is significant for all three base models: $+6.9$, $+9.7$ and $+4.6$ points. It is smallest for Qwen3-235B, whose draft already recovers the most in-transcript content ($64.1$), so it has the least room to improve. Beyond-transcript gains are small ($+2.1$, $+2.0$, $+0.8$) and only one is significant. Restricting to the unanimous subset sharpens the contrast: the in-transcript gain holds for two of the three base models, while every beyond-transcript gain falls to $+1.3$ or less with $p\ge0.317$.

\begin{table}[h]
\centering
\caption{FactEHR claim recall split by whether the transcript entails the gold claim. The $n$ columns partition each base model's gold claims into
the two strata; $p$-values are Wilcoxon signed-rank over encounters. All three base models share
the same 862 gold claims, except Qwen2.5-72B, scored over 819 because its revision looped on two
encounters that the judge excludes.}
  
\label{tab:stratified_consensus}
\footnotesize
\setlength{\tabcolsep}{3.5pt}
\begin{tabular}{llcccccccc}
\toprule
 & & \multicolumn{4}{c}{\textbf{in-transcript}} & \multicolumn{4}{c}{\textbf{beyond-transcript}} \\
\cmidrule(lr){3-6}\cmidrule(lr){7-10}
\textbf{Model} & \textbf{Label} & $n$ & Draft & $+$CDR & $\Delta$ ($p$) & $n$ & Draft & $+$CDR & $\Delta$ ($p$) \\
\midrule
\multirow{2}{*}{Qwen2.5-32B}
 & Majority ($\geq$3 of 5)  & 387 & 45.0 & 51.9 & $+6.9$ (0.000) & 475 & 8.4  & 10.5 & $+2.1$ (0.010) \\
 & Unanimous (5 of 5) & 156 & 61.5 & 67.9 & $+6.4$ (0.017) & 225 & 2.7  & 3.1  & $+0.4$ (0.317) \\
\addlinespace[2pt]
\multirow{2}{*}{Qwen2.5-72B}
 & Majority ($\geq$3 of 5)  & 373 & 50.4 & 60.1 & $\mathbf{+9.7}$ (0.000) & 446 & 9.0 & 11.0 & $+2.0$ (0.064) \\
 & Unanimous (5 of 5) & 149 & 66.4 & 75.2 & $+8.8$ (0.014) & 214 & 5.6 & 5.6  & $+0.0$ (1.000) \\
\addlinespace[2pt]
\multirow{2}{*}{Qwen3-235B}
 & Majority ($\geq$3 of 5)  & 387 & 64.1 & 68.7 & $+4.6$ (0.022) & 475 & 17.3 & 18.1 & $+0.8$ (0.417) \\
 & Unanimous (5 of 5) & 156 & 80.8 & 84.6 & $+3.8$ (0.113) & 225 & 7.6  & 8.9  & $+1.3$ (0.916) \\
\bottomrule
\end{tabular}
\end{table}

The split also rules out a simpler explanation of the gain. CDR only adds text, so a longer note
could match more clinician claims by chance. But the strata are defined by the transcript, not
by the note, so added text has no way to favor one over the other: an effect of length alone
would show up equally on both. It does not. The in-transcript gain is $3$--$6\times$ the
beyond-transcript gain, and two of the three beyond-transcript changes are not significant.

\section*{Discussion}

\noindent\textbf{The full picture: precision and F1.}
CDR only adds text, so a rise in recall can come with a fall in precision. It does, in every
condition. On the SOAP-prompt base on Pitt-Bench, averaged over the four base models, MedCon
precision falls from $12.74$ to $11.28$ and F1 from $18.82$ to $17.46$. FactEHR moves the same
way: precision from $26.29$ to $21.96$, F1 from $26.75$ to $25.44$. Recall rises in every one of
those cases (Table~\ref{tab:main_recall}).
The trade comes from the short reference, not from the transcripts. Scored against a
median-881-character therapist note, a longer and more complete note is penalised, and we showed
earlier that most of the added claims are in fact supported by the transcript. This is the
concrete reason we treat recall as the primary endpoint and report these numbers for
transparency rather than as a ranking.

\noindent\textbf{CDR is more than a second look.} CDR asks the reviser to look at the
draft a second time, so part of any gain could come from the second pass alone rather than from
the evidence CDR supplies. We test this with a control that changes only the evidence: the same
draft, the same reviser, and the same mandatory-add instruction, but the full transcript in
place of the missing concepts CDR identifies. CDR recovers more of the therapist's concepts for
all three base models, improving MedCon recall by about $5\%$ in relative terms, and the two
arms reach different content: $87$ to $93\%$ of what CDR adds is never reached by the transcript
control.

\noindent\textbf{What works.} A graph built from the transcript is a usable checklist for what
a draft left out. Recall rises in every condition we test under both metrics, and the gain
lands where the method can reach: on claims the transcript entails, not on the rest. The
\emph{reviser} need not be the base model, nor a frontier one, since an open-weight local model
revising the GPT-5 Pro draft is competitive with GPT-5 Pro revising its own. The judges run
locally too, so the whole pipeline can sit inside institutional infrastructure.

\noindent\textbf{What it cannot fix.} Roughly half of what a therapist writes never reaches the
recording. Gait quality, assistance level and transfer performance are observed and measured,
not spoken. No transcript-based system can recover them however good the graph, which is why
our beyond-transcript gains are small and mostly not significant. This caps the approach and
points to video or sensor input rather than to a better language model.

\section*{Limitations}
Pitt-Bench is a single site with few speech-language sessions, and we generate once per
encounter, so run-to-run variation is not measured. Claim decomposition and entailment are
LLM-based and will shift with model version. ``In transcript'' means entailed by the
transcript, not literally spoken.
MedCon's semantic-type filter excludes most rehabilitation
activity content, so its absolute values understate agreement on this corpus. One detail bounds
the GPT-5 Pro rows in particular: their graphs are built by Qwen3-235B rather than by GPT-5
Pro, because API cost makes GPT-5 Pro graph construction impractical on full-length therapy
transcripts, so those rows differ from the others in \emph{reviser} and in graph source at once.

\section*{Conclusion}
Coverage-Directed Revision recovers clinician-documented content the base model dropped,
without touching the ambient AI model, so it attaches to a scribe already in use. The gain is
concentrated in content the transcript supports and holds across every base, \emph{reviser} and judge
we test, though it amounts to about one recovered claim per note. The checklist is only as good
as the graph behind it, so a cleaner graph --- better extraction, tighter grounding, less
non-clinical noise --- is the most direct route to a stronger draft and to improving precision
and F1 alongside recall. What remains missing is mostly information that never reached the
recording at all, which bounds what any audio-only system can do and points to multimodal
capture as the next step.

\subparagraph{Acknowledgments}

The research reported in this article was supported by the National Institutes of Health awards R01HD117897 and R01LM014588. The sponsors had no role in study design, data collection, analysis, interpretation, report writing, or decision to submit the paper for publication.

\makeatletter
\renewcommand{\@biblabel}[1]{\hfill #1.}
\makeatother

\renewcommand{\refname}{\centerline{References}}

\bibliographystyle{vancouver}
\bibliography{reference}

\end{document}